\documentclass[a4paper]{article}
\usepackage{INTERSPEECH_v2}
\usepackage{graphicx}
\usepackage{multirow}
\usepackage{amsmath}

\title{EFFECT OF DATA REDUCTION ON SEQUENCE-TO-SEQUENCE NEURAL TTS}
\name{Javier Latorre, Jakub Lachowicz, Jaime Lorenzo-Trueba, Thomas Merritt, Thomas Drugman, Srikanth Ronanki, Klimkov Viacheslav,  }
\address{Amazon.com}
\email{\{jlatorre, lachj, truebaj, thommer, drugman, ronanks, vklimkov \}@amazon.com}

\begin{document}

\maketitle
\makeatletter
\def\blfootnote{\gdef\@thefnmark{}\@footnotetext}
\makeatother
\blfootnote{Paper submitted to IEEE ICASSP 2019}
\begin{abstract}
Recent speech synthesis systems based on sampling from autoregressive neural networks models can generate speech almost undistinguishable from human recordings.
However, these models require large amounts of data.   
This paper shows that the lack of data from one speaker can be compensated with data from other speakers.
The naturalness of Tacotron2-like models trained on a blend of 5k utterances from 7 speakers is better than that of speaker dependent models trained on 15k utterances, but in terms of stability multi-speaker models are always more stable.
We also demonstrate that models mixing only 1250 utterances from a target speaker with 5k utterances from another 6 speakers can produce significantly better quality than state-of-the-art DNN-guided unit selection systems trained on more than 10 times the data from the target speaker.
\end{abstract}

\section{Introduction}
Data acquisition is one of the main problems of data-driven text-to-speech (TTS) systems.  
High-quality  unit selection TTS relies on large  single speaker databases, usually of tens of hours of speech.
Classical statistical parametric speech synthesis (SPSS) is more data frugal. 
Less than one hour of data is enough to train an intelligible speaker dependent (SD) model. 
More data improves SPSS quality, but from  around 4-5 hours of data onwards quality tends to saturate \cite{HMM_robustness}. 
To reduce the dependency on a single speaker,  techniques based on mixing data from multiple speakers into an Average Voice Model (AVM) were developed. 
These techniques produce reasonable quality with as little as 3 minutes of target speaker data \cite{yamagishi_AVM_Review}.
However, when the available target speaker data is above 2 hours ($\sim$2k utterances), Speaker-Dependent (SD) models were better \cite{thousandVoices}.  
 
The change of paradigm introduced by  auto-regressive models \cite{oord2016wavenet, ping2017deep3, arik2017deep, Wang2017, tacotron2}, has produced synthetic speech of unprecedented quality. 
These new models require much more data than traditional TTS but they are also more efficient at integrating diverse data \cite{arik2017deep2, lorenzo2018investigating, Podsiad?o2018}. 
Several studies have reported that it is easy to train multi-speaker models \cite{arik2017deep2, Jia2018TransferLF} and that adding more speakers improves the loss function over the validation set \cite{oord2016wavenet}.
Most approaches for multi-speaker models rely on a speaker embedding but they vary on the type of embedding and where to apply it. 
Whereas some use an external model, e.g. speaker classification, to provide the embeddings \cite{VoiceLoop, Jia2018TransferLF} others train the speaker embedding together with the model out of a one-hot speaker ID vector \cite{oord2016wavenet, arik2017deep2, ping2017deep3}.
Some approaches use the embedding at the input only as a global conditioning \cite{oord2016wavenet}, whereas others apply it at different levels within the model \cite{arik2017deep2, ping2017deep3}.

Despite all the recent attention to multi-speaker models, to the best of our knowledge, nobody has published yet any study on practical issues such as; 1) at which point an SD model becomes better than a multi-speaker one,  2)  whether it is better or worse to use gender-dependent multi-speaker models, 3)  what is the effect of training models with an  unbalanced mixture of data from the target speaker and other speakers. 
This paper presents the results of several experiments aimed at answering these questions. We hope our results will help other developers and researchers in designing their systems and experiments.


The structure of the paper is as follows: 
Section~\ref{sec:our_system} describes the basic structure of our TTS system; 
Sections~\ref{sec:experiments} and~\ref{sec:results} describe the experimental protocol and results respectively. 
Finally, in Section~\ref{sec:conclusions} conclusion are drawn.



\section{System description} \label{sec:our_system}
Our system  architecture follows that of Tacotron2 \cite{tacotron2}. First, a sequence-to-sequence (S2S) acoustic model predicts the mel-spectrograms from a sequence of linguistic inputs. Then a neural vocoder converts the mel-spectrograms into a waveform.  

\subsection{Acoustic model}
 
\begin{figure}[tb!]
\centering
\includegraphics[width=80mm]{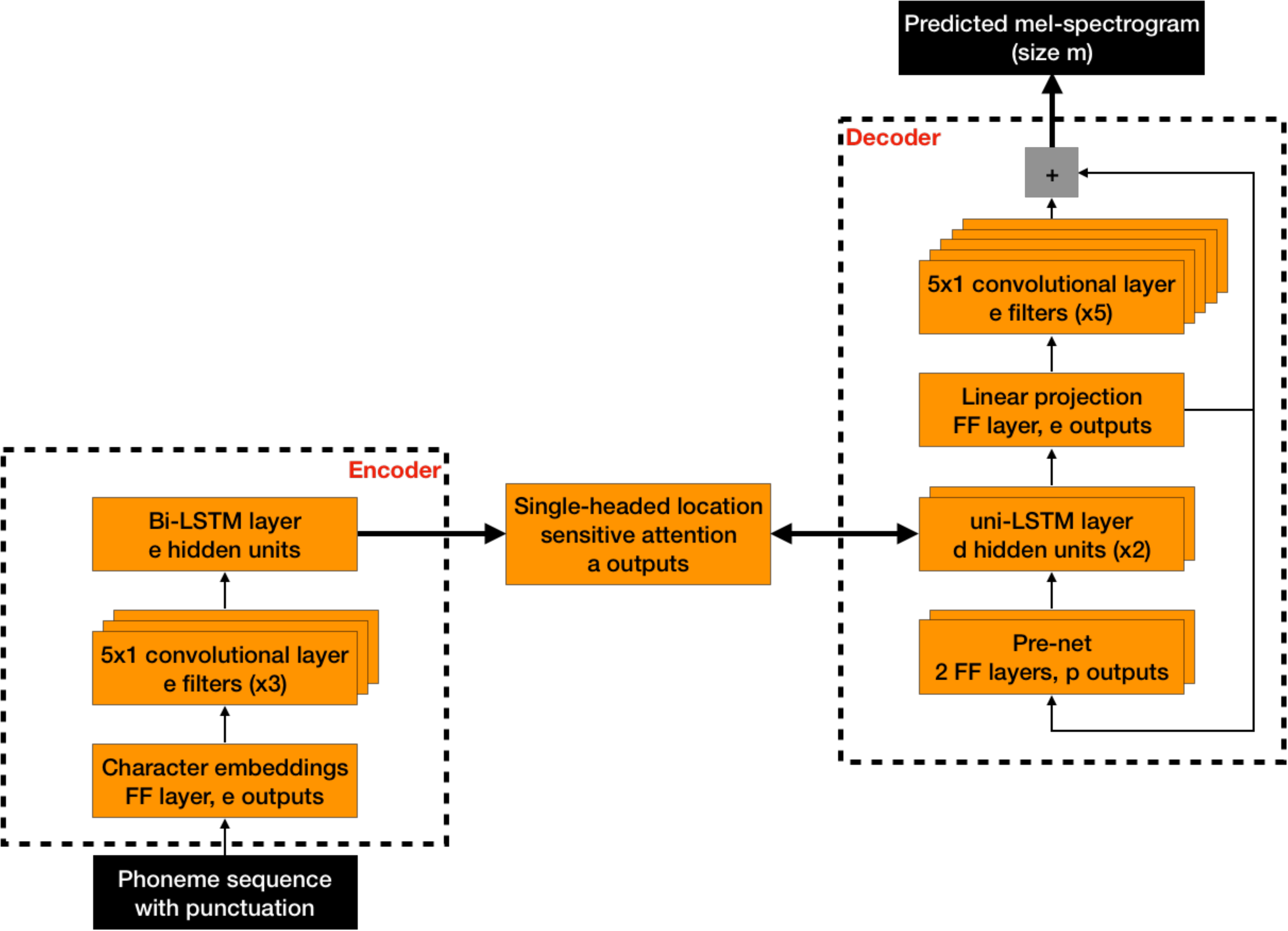}
\caption{Acoustic model architecture}\label{fig:prosotron_topology}
\vspace{-2mm}
\end{figure}

The architecture of the acoustic model is described in figure~\ref{fig:prosotron_topology}. It is a S2S model with attention mechanism as in \cite{tacotron2}. However, instead of using raw graphemes as inputs, our system first converts the graphemes into phonemes which are then encoded with a one-hot vector. For the vowels, we use 3 different symbols depending on their level of stress (0,1,2).
The punctuation after each word, including blanks, is treated as if it were another phoneme.  

The attention mechanism for the S2S model follows the one proposed in \cite{BahdanauCB14} with normalised attention weights \cite{SalimansK16}. 
In this mechanism the  attention weights for the current frame depend both on the previous output of the decoder and on the attention weights of the previous frame. 
The speaker conditioning is similar to  \cite{oord2016wavenet},  with a one-hot speaker ID and global conditioning.    

The output of the model are blocks of 5 frames of mel-spectrograms, each consisting of an 80-dimensional vector spanning frequencies between 50 Hz and 12 kHz.  Each frame is computed over 50 ms and shifted every 12.5 ms. 
The last frame of the previous block is passed as input to  both the attention model and the decoder to generate the next 5-frame block. 
During training, this recursive input is randomly switched between real spectrograms and self-generated ones (scheduled sampling). The probability for taking real spectrograms is 0.9.
In addition to the mel-spectrograms, the model also predicts a stop token to mark the end of the utterance. The stop token is encoded as a real number between 0 and 1 that reaches the value of 1 at the end of the sentence.
The model was trained with a dropout probability of 0.1 for both the decoder and the auto-regression, but without dropout for the encoder.  The dropout was also applied at inference time.

\subsection{Neural vocoder}
The architecture of the neural vocoder closely follows WaveRNN \cite{kalchbrenner2018efficient}.  The autoregressive part of the network consists of a single forward Gated Recurrent Unit with a hidden size of 896 and a pair of affine layers followed by a softmax layer with 1024 outputs to predict the 10-bit mu-law samples with 24 kHz sampling rate. The conditioning network consists of a two bi-directional Long Short-Term Memory (LSTMs) with a hidden size of 128. The mel-spectrograms for conditioning  consisted of 80 coefficients extracted using Librosa library \cite{mcfee2015librosa} for frequencies from 50 Hz to 12 kHz.
The model was trained on data from 74 speakers on 17 different languages with between 1k to  2.5k utterances per speaker. 
 Around two thirds of the speakers were female and the other third male, except for one child.  
 More details about the vocoder architecture and how it was trained can be found in \cite{Lorenzo2019ICASSP}.
 
\section{Experimental protocol}\label{sec:experiments}
The  research questions we attempted to answer were:
\begin{enumerate}
\item Can a multi-speaker model with limited data per speaker achieve similar quality than a SPSS guided unit selection with a large database? 
\item Can we train multi-speaker models with less data for the target speaker than for the supporting speakers?
\item How much data is needed for a  SD model to be better than a multi-speaker one?
\item Is it better to combine all the available speakers or only the most similar ones, e.g., only female speakers?
\item Does mixing speakers affects the speaker similarity?
\end{enumerate}
The results of our  experiments to answer these questions are presented in section~\ref{sec:results}.
\subsection{Training data and model stability}\label{sec:stability}
The speech data used to train the models came from 7 internal speakers: 2 males, 4 females and one child. The available data for these speaker was 8.5k utterances for four speakers (2 female, one male and the child), 15k for two (one male and one female) and 25k utterances for one female speaker. 
Out of this, we randomly selected a fixed number of utterances per speaker depending on the model. For each speaker in the model, we used 90\% of the utterances for training and 10\% for development.  
The first three columns of table~\ref{table:Stability} shows the speakers blend and amount of data used to train each type of model in our experiments. 

A problem in S2S models is that the attention sometimes gets lost at inference time. 
This produces errors such as skipping one or more phones, repeating part of the sentence, getting stuck in silences, etc.  
An analysis of the stability of the models is useful to understand their robustness toward different blends of training data. 
To measure this, we generated 75 utterances from each speaker on each type of model and marked  those that, after listening, presented any of the above mentioned stability problems. 
The last column of table~\ref{table:Stability} shows the proportion of stable utterances for each model. 
SD models are clearly much more unstable than multi-speaker one, regardless of whether the mixed speaker ones are female-only  or mix-gender. 
This result agrees with the comments on \cite{oord2016wavenet} about convergence of multi-speaker models.
Model stability does not seem to be directly linked to amount of training data. 
The female-only model trained on 2.5k utterances/speaker was more stable than the female-only models trained on more data.  
Also,  some multi-speaker models are more stable than SD ones, despite being trained on less data.
The type of problems seem to depend on the speaker, even in the multi-speaker models. 
All these suggest that stability depends on the characteristics of the data itself, but we could not find any clear pattern for it. 

\begin{table}
 \caption{Percentage of correctly generated files }\label{table:Stability}
\begin{tabular}{|c|c|c|c|}
  \hline
model & model name& \#training utt   &\% stable\\
\hline
single speaker & sd-8500&8.5k & 35.4\%\\
(1 speaker)			& sd-15000 &15k & 46.2\%\\
						& sd-25000 &25k & 69.3\%\\
						\hline
female only& fe4-2500&4$\times$2.5k &  88.3\%\\
		 (4 speakers) 			  & fe4-5000&4$\times$5k & 77.33\% \\
				          & fe4-8500&4$\times$8.5k &  77.33\% \\
						
\hline
mix-gender 	& mx7-2500& 7$\times$2.5k &  54.5 \%\\
(7 speakers)	& mx7-5000& 7$\times$5k & 93.5\%\\
			& mx7-8500& 7$\times$8.5k & 95.6\%\\
						\hline
mix-gender unbalanced & mx6+1250& 6$\times$5k $+$ 1.25k & 91.4\%\\
(7 speakers)	             &mx6+2500& 6$\times$5k $+$ 2.5k &  78.9\%\\
						
 \hline
\end{tabular}
\vspace{-4mm}
\end{table}

\subsection{Subjective evaluation}
To address our research questions, we ran several MUltiple Stimuli with Hidden Reference and Anchor (MUSHRA) tests \cite{recommendation2001bs}.
The advantage of MUSHRA over Mean Opinion Score (MOS) is that for each sentence, all the systems being evaluated can be presented simultaneously in one panel. 
On every test the positions of the systems on the panel were randomised. 
All the panels included the natural recordings as an upper anchor but similar to  \cite{AmawaveEval} subjects were not forced to assign the top score to any of the systems. 
All tests were conducted in Amazon Mechanical Turk. Subjects were people living in the United States who define themselves as native English speakers. 
For each evaluation, we selected sentences of length between 5 and 30 words.
For all the tests the significance of the results was analysed with a Wilcoxon signed-rank test and a standard t-test, both with Bonferroni-Holm correction applied \cite{WilcoxonRob}.
The main goal of the subjective tests was to evaluate speech quality and speaker similarity. Therefore for each of the MUSHRA tests we chose only those utterances that did not present any stability problem with any of the systems under consideration.
\subsubsection{Naturalness}
For tests on questions 1-4 subjects were asked  to ``rate the audio samples in terms of their naturalness"  with a continuous slider between  ``completely unnatural" (0) and ``completely natural" (100).   Each stimuli panel was evaluated by 10 subjects.  
The set of sentences used on each experiment were slightly different since due to stability problems, not all the systems on each test were able to synthesise all the utterances.
In the tests for questions 1 and 2, a guided unit selection was included among the systems to be evaluated. This guided unit selection was a standard system in which the linguistic cost is combined with the acoustic cost, computed as the distance between the F0, duration and spectrum of the units and those  predicted by a state-level DNN model. The models for the acoustic cost were speaker dependent and trained with all the available data for each speaker. At synthesis time, the evaluation sentences were blacklisted so that their units could not be selected. This blacklisting removed less than 0.5\% of the unit selection data. Since those tests proved that the guided unit selection was worse than the other systems, we didn't include it for the rest of the experiments.
\subsubsection{Speaker similarity}
For the test on question 5 subjects were asked to ``rate whether the speaker of the reference sounds like the same person as the speakers of the samples." between ``Definitely a different person'' (0) and ``Definitely the same person'' (100). 
Subjects were presented with a reference audio from the target speaker (sentence1)  and speech audio samples for a different sentence (sentence2) generated by the evaluated models. 
The recording of sentence2 by the target speaker was also included as an upper anchor. 
For each of the seven speakers, we ran an independent MUSHRA test with 10 utterances and the best available SD model for that speaker.


\section{Results}\label{sec:results}

\subsection{Multi-speaker vs unit selection}\label{sssec:mixture}
The first experiment evaluates the naturalness of two multi-speaker models, `mx7-5000' and `mx7-2500'  (see table~\ref{table:Stability}) vs the guided unit selection.
As an additional reference point, we included samples re-synthesised from the original mel-spectrograms with the neural vocoder, `nv-resynthesis'. 

\begin{figure}[t] 
\centering
	\includegraphics[width=1\linewidth]{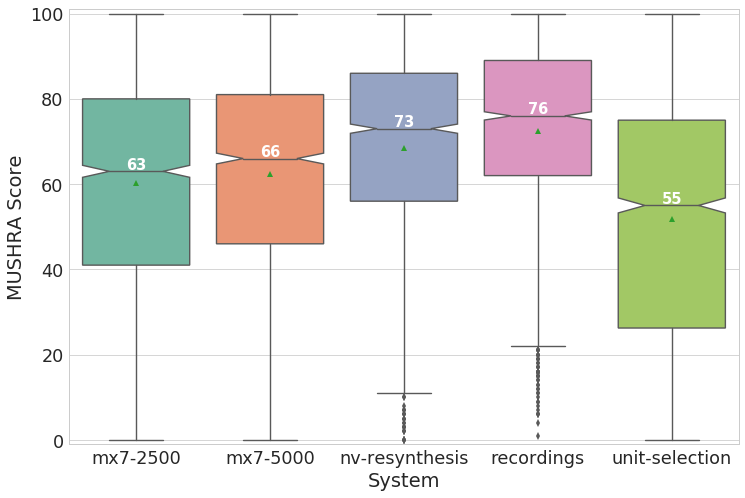}
\caption{{ Multi-speaker models vs. Unit selection}}\label{fig:2500vs5000_MUSHRA}
\vspace{-2mm}
\end{figure}

The evaluation consisted of 27 utterances from each of the 7 speakers resulting in a total of 189 stimuli panels. A total of 70 subjects evaluated 27 panels each. 
The boxplots of the MUSHRA scores can be seen in figure~\ref{fig:2500vs5000_MUSHRA}.
All the models were significantly different from each other at  $p<0.05$. 
As expected, the recordings and the `nv-resynthesis' samples achieved the higher score followed by the `mx7-5000' and `mx7-2500'. 
The difference between the `mx7-2500' and `mx7-5000' is small but statistically significant. 
The most surprising result was the comparatively low score of the guided unit selection, despite it being built upon more than 99\% of all the available data.
Obviously, there were differences between speakers, but they do not correlate with the amount of data of the unit selection system. The rank order of the systems was consistent across speakers. 
The median MUSHRA in figure~\ref{fig:2500vs5000_MUSHRA} also shows that the gap between `nv-resynthesis' and the recordings is very small, despite the vocoder being a generic one trained on multiple  speakers in different languages. The main gap is between the models and the `nv-resynthesis', i.e. in the modelling of the mel-spectrograms. 
Comparatively,  the gap due to difference in the amount of training data is smaller.


\subsection{Balanced vs unbalanced mixture of speakers}
The second experiment evaluated the naturalness of the models of previous experiments vs models trained with 5k utterances from six speakers plus 2.5k or 1.25k utterances from a target speaker, `mx6+2500' and `mx6+1250' respectively. We train one `mx6+...' model for each speaker and used them only to generate speech with the voice of that speaker.   
To keep the lower anchor of the previous experiment (section~\ref{sssec:mixture}), we added again samples produced by the guided unit selection system. 
 
\begin{figure}[t] 
\centering
\includegraphics[width=1\linewidth]{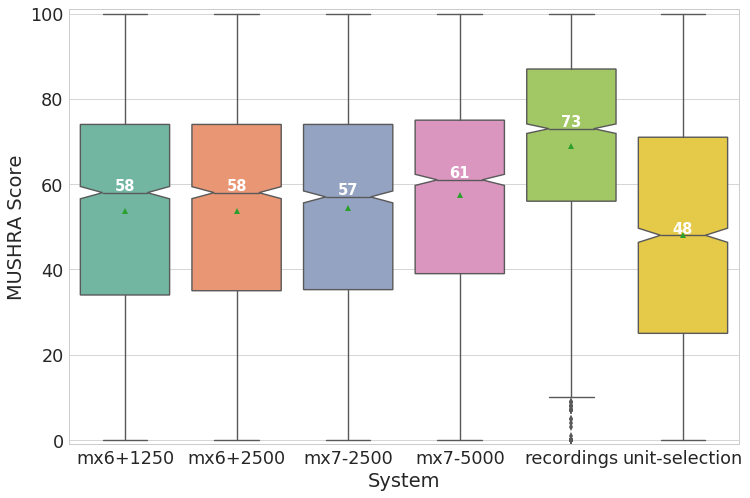}
\caption{{Mixed models with balanced vs unbalanced data}}\label{fig:balanceVunbalance_MUSHRA}
\vspace{-5mm}
\end{figure}

The evaluation consisted of 27 utterances from each of the 7 speakers. They were evaluated by a total of 70 subjects. 
Figure~\ref{fig:balanceVunbalance_MUSHRA} shows the results.  
The ranks of the results of `unit-selection', `mx7-2500', `mx7-5000' and `recordings' confirm the results of Section \ref{sssec:mixture}. 
The `mx7-2500', `mx6+1250' and `mx6+2500' are not significantly different from each other. 
This indicates that the benefit of using 5k utterances instead of 2.5k  for the non-target speakers is not in terms of quality, 
but in terms of stability as was shown in section~\ref{sec:stability}.  
A second interesting result was that in terms of quality, 1250 utterances from a target speaker mixed with sufficient data from other speakers can generate better speech quality than a state-of-the-art unit selection system. The only exception to this was the female speakers on $>$15k utterances.
There were some other minor differences between speakers, especially in the relative ranking of the two `mx6' models. However, with the above mentioned exception, the rank order between systems was  consistent across speakers. 

%

\subsection{Multi-speaker vs speaker dependent}
This set of experiments compared SD models with multi-speaker models `mx7-5000'  and `mx7-8500' (see table~\ref{table:Stability}).
We trained `sd-8500' models for all seven speakers, `sd-15000'  for 3 speakers and `sd-25000' for one speaker, depending on the amount of data available for the speakers. 
Three separated evaluations were conducted, for the SD models on 8.5k, 15k and 25k utterances.

Unfortunately, out of the seven `sd-8500' models only 3 (two female and one male) were stable  enough to generate samples. To compensate for the lack of data points in the evaluation of the `sd-8500' models we used 42 samples for each of the remaining 3 speakers.  
For the evaluation of the `sd-15000' models we only have 3 speakers with enough data. As shown in table~\ref{table:Stability} these models were also very unstable, especially for one of the speakers which only generated correctly 24 utterances. To keep the number of utterances per speaker balanced we evaluated these models with 24 utterances/speaker.  
Finally, for the `sd-25000' we generated 45 sentences from that speaker model. To compensate for the lack of data, each MUSHRA panel of the `sd-25000' evaluation was judged by 15 subjects.

Table~\ref{table:sd-MUSHRA-score} shows the median MUSHRA score for the three tests and the average rank of the systems . 
In the three evaluations, the difference between `mx7-5000' and `mx7-8500' were not statistically significantly as can be seen by the small differences in their averaged rank order. 
Both multi-speaker models were better than `sd-8500' models and worse than the  `sd-25000' model.  The `mx7-8500' model was better than the 'sd-15000' model. 
These differences were statistically significant.  
The differences between `sd-15000' model and `mx7-5000' were not significant with the Wilcoxon signed-rank test. 

These results suggest that, similar to classical SPSS, a SD model can sound more natural than a multi-speaker one when trained on sufficient amount of data. 
However, multi-speaker models are better than SD models when they are trained on more than 2.3 times more data or, alternatively, when the SD model is trained on less than 15 hours. Further work is needed to clarify this last point.
%
\vspace{-1mm}
\begin{table}[htbp] 
 \caption{Median score and average rank (in parentheses) of the tests comparing multi-speaker vs speaker dependent models }\label{table:sd-MUSHRA-score}
\begin{tabular}{|c|c|c|c|c|}
  \hline
Evaluated  & Recordings& \multicolumn{3}{|c|}{Models}\\
\cline{3-5}
 SD model& &SD & mx7-8500 & mx7-5000\\
\hline
sd-8500 & 71 (1.96) & 61 (2.78) & 63 (2.61) & 62 (2.64) \\ 
\hline
sd-15000 & 74 (1.91) & 61.5 (2.79) & 63 (2.65) & 62 (2.65) \\
\hline
sd-25000  & 77 (1.97) & 68 (2.56) & 67 (2.73) & 66 (2.75)\\
\hline
\end{tabular}
\vspace{-2mm}
\end{table}
 

\vspace{-4mm}
\subsection{Female only  vs mixed gender}
\begin{figure}[htbp] 
\vspace{-8mm}
\centering
\includegraphics[width=1\linewidth]{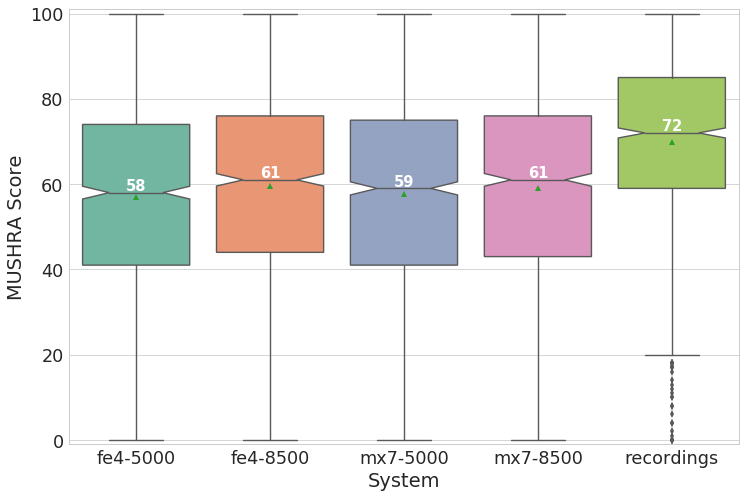}
\caption{{Mixed of all speakers vs only female speakers}}\label{fig:gender_dependent._MUSHRA}
\vspace{-4mm}
\end{figure}

The last naturalness experiment compared models trained on all 7 speakers against those trained only on 4 female speakers. The total amount of data was different but the amount of data per speaker was constant.
Figure~\ref{fig:gender_dependent._MUSHRA} shows the results. 
The differences between models trained  on different number of utterances per speaker were statistically significant but  the differences between models trained on the same data per speaker were not. 

\subsection{Speaker similarity}
Table~\ref{table:speaker-similarity} summarises the results. Each row represents a different speaker. 
On the average, only the differences between recordings and the other systems were statistically significant at $p<0.05$. 
On a per-speaker basis, the differences for two speakers (in bold type face in table~\ref{table:speaker-similarity} ) and the rest of the models were statistically significant. 
However, that significancy dissappears when both 8500 or 15000 speakers are considered jointly.
\begin{table}[htbp] 
 \caption{Average speaker similarity }\label{table:speaker-similarity}
\begin{tabular}{|c|c|c|c|c|c|c|c|}
  \hline
Evaluated  & Record- & \multicolumn{6}{|c|}{Models}\\
\cline{3-8}
SD model &ings & best &  \multicolumn{3}{|c|}{mx7-...} & \multicolumn{2}{|c|}{mx6+...}\\
\cline{4-8}
 & & SD &8500& 5000 & 2500 & 1250 & 2500\\
\hline
\multirow{4}{*}{sd-8500}& 78.3& 69.5& 70.7& 70.2& 68.8& 70.0& 70.1 \\ 
\cline{2-8}
& \bf{73.0}& {\bf 68.1}& {\bf 76.3}& {\bf 74.7}& {\bf 76.7}& {\bf 71.9}& {\bf 73.0}\\ 
\cline{2-8}
& 76.5& - & 70.7& 71.3& 71.0& 71.3 & 71.4 \\ 
\cline{2-8}
& 79.3& - & 71.5& 73.3& 74.6& 69.1 & 73.2\\ 
\hline
\multirow{2}{*}{sd-15000}& 68.1& 68.4& 65.2& 68.2& 67.4& 68.4& 64.5\\ 
\cline{2-8}
&{\bf 83.4} & {\bf 77.3}& {\bf 82.3}& {\bf 84.3}& {\bf 84.2}& {\bf 82.3}& {\bf 82.3} \\ 
\hline
sd-25000& 75.0& 72.1& 69.9& 71.4& 70.7& 71.8& 70.1\\ 
\hline
Average & 76.0& 70.7& 72.1& 73.1& 73.2 & 71.3& 72.2\\ 
\hline
\end{tabular}
\vspace{-2mm}
\end{table}
\vspace{-2mm}
\section{Conclusions}\label{sec:conclusions}
This paper presents several experiments aimed at reducing the amount of single speaker data needed to train high-quality S2S TTS systems. 
The results show that models trained on a mixture of speakers can produce better quality than a state-of-the-art guided unit selection TTS with an inventory of units ranging between 8.5k and 25k utterances.
We show that this is true for S2S models trained on 2.5k utterances from 7 speakers and also for S2S models trained on a mixture of 1.25k utterances from the target speaker and 5k utterances from 6 other speakers. 
Our results also show that for databases with up to 15k utterances, multi-speaker models produce better quality than speaker-dependent ones. SD models with more data can produce marginally better quality but in terms of stability SD models are always more unstable.
The most probable reason for this is that by mixing multiple speakers, the alignment is more robust against different pronunciations,  wrong sentences or different initialisation values.  
This seems to also be the case when training on less data but more similar speakers. The different speaker blends do not seem to affect the speech quality but lower variability of speaker seems to impact negatively on the model's stability.  


\clearpage
\newpage
\bibliographystyle{IEEEtran}

\bibliography{references}

\end{document}